\documentclass[12pt]{article}

\usepackage{times}
\usepackage{graphicx}
\usepackage{hyperref}
\usepackage{url}

\title{The Shape of Art History in the Eyes of the Machine\footnote{This paper is an extended version of a paper that will be published on the 32nd  AAAI conference on Artificial Intelligence, to be held in New Orleans, USA, February 2-7, 2018}}

\author{\href{mailto:elgammal@cs.rutgers.edu}{Ahmed Elgammal}$^1$\footnote{
Corresponding author: {Ahmed Elgammal \href{mailto:elgammal@cs.rutgers.edu}{elgammal@cs.rutgers.edu}}}     \and
Marian Mazzone$^2$ \and
 Bingchen  Liu$^1$ \and Diana Kim$^1$ \and Mohamed Elhoseiny$^1$
\\
\href{http://digihumanlab.rutgers.edu}{The Art and Artificial Intelligence Laboratory}  \\
$^1$ Department of Computer Science, Rutgers University, NJ, USA  \\
$^2$ Department of Art History, College of Charleston, SC, USA
}

\date{}

\graphicspath{{../}}

\begin{document} 
\maketitle

\begin{abstract}

How does the machine classify styles in art? And how does it relate to art historians' methods for analyzing style? Several studies have shown the ability of the machine to learn and predict style categories, such as Renaissance, Baroque, Impressionism, etc., from images of paintings. This implies that the machine can learn an internal representation encoding discriminative features through its visual analysis. However, such a representation is not necessarily interpretable. We conducted a comprehensive study of several of the state-of-the-art convolutional neural networks applied to the task of style classification on 77K images of paintings, and analyzed the learned representation through correlation analysis with concepts derived from art history. Surprisingly, the networks could place the works of art in a smooth temporal arrangement mainly based on learning style labels, without any a priori knowledge of time of creation, the historical time and context of styles, or relations between styles. The learned representations showed that there are few underlying factors that explain the visual variations of style in art. Some of these factors were found to correlate with style patterns suggested by Heinrich W\"olfflin (1846-1945). The learned representations also consistently highlighted certain artists as the extreme distinctive representative of their styles, which quantitatively confirms art historian observations.

\end{abstract}
 
\pagebreak

\begin{center}
{Art history could prove ``to contain unexpected potentialities as a predictive science'' -- George Kubler, The Shape of Time: Remarks on the History of Things, 1962.}
\vspace{6pt}
\end{center}

\section{Introduction}
Style is central to the discipline of art history. The word ``style'' is used to refer to the individual way or manner that someone makes or does something, for example Rembrandt's style of painting. Style also refers to groups of works that have a similar typology of characteristics, such as the Impressionist style, or High Renaissance style.  Art historians identify, characterize, and define styles based on the evidence of the physical work itself, in combination with an analysis of the cultural and historical features of the time and place in which it was made.   Although we see style, and we all know that it exists, there is still no central, agreed upon theory of how style comes about, or how and why it changes.  Some of the best scholars of art history have written persuasively about the importance of style to the discipline, and the concomitant difficulty of defining or explaining what it is and why it changes \cite{1,2}. Up to now connoisseurship has proven to be one of the most effective means to detect the styles of various artists, and differentiate style categories and distinctions in larger movements and periods.

Recent research in computer vision and machine learning have shown the ability of the machine to learn to discriminate between different style categories such as Renaissance, Baroque, Impressionism, Cubism, etc., with reasonable accuracy, (e.g. \cite{3,4,5,6}). However, classifying style by the machine is not what interests art historians. Instead, the important issues are what machine learning may tell us about how the characteristics of style are identified, and the patterns or sequence of style changes. The ability of the machine to classify styles implies that the machine has learned an internal representation that encodes discriminative features through its visual analysis of the paintings. However, it is typical that the machine uses visual features that are not interpretable by humans. This limits the ability to discover knowledge out of these results. 

Our study's emphasis is on understanding how the machine achieves classification of style, what internal representation it uses to achieve this task, and how that representation is related to art history methodologies for identifying styles. To achieve such understanding, we utilized one of the key formulations of style pattern and style change in art history, the theory of Heinrich W\"olfflin (1846-1945). W\"olfflin's comparative approach to formal analysis has become a standard method of art history pedagogy. W\"olfflin chose to separate form analysis from discussions of subject matter and expression, focusing on the ``visual schema'' of the works, and how the ``visible world crystallized for the eye in certain forms'' \cite{7}. W\"olfflin identified pairs of works of art to demonstrate style differences through comparison and contrast exercises that focused on key principles or features. W\"olfflin used his method to differentiate the Renaissance from the Baroque style through five key visual principles: linear/painterly, planar/recessional, closed form/open form, multiplicity/unity, absolute clarity/relative clarity. W\"olfflin posited that form change has some pattern of differentiation, such that style types and changes can only come into being in certain sequences.

With advances in computer vision and machine learning and the availability of comprehensive datasets of images, we are now positioned to approach the history of art as a predictive science, and relate its means of determining questions of style to machine results. Lev Manovich has argued that the use of computational methods, while providing a radical shift in scale, in fact continues the humanities' traditional methodologies \cite{8}. It was nearly impossible to apply and empirically test W\"olfflin's methods of style differentiation and analysis before developments in computer science.  No human being would assemble the number of examples needed to prove the value of his methods for finding discriminative features. Nor could anyone amass the dataset necessary to demonstrate the usefulness of his model for processing style description and sequencing beyond the immediate examples of Renaissance and Baroque via his five pairs.  We chose W\"olfflin's theory because of his emphasis on formal, discriminative features and the compare/contrast logic of his system, qualities that make it conducive to machine learning. Today, art historians use a wide variety of methods that are not only solely focused on form, but for the type of analysis of this paper W\"olfflin's approach is useful.


\subsection*{Methodology}
Deep convolutional neural networks have recently played a transformative role in advancing artificial intelligence \cite{9}. We evaluated a large number of state-of-the-art deep convolutional neural network models, and variants of them, trained to classify styles. We focused on increasing the interpretability of the learned presentation by forcing the machine to achieve classification with a reduced number of variables without sacrificing classification accuracy. We then analyzed the achieved representations through linear and nonlinear dimensionality reduction of the activation space, visualization, and correlation analysis with time and with W\"olfflin's pairs. We used a collection of 77K digitized paintings to train, validate and test the models. We utilized two sets of digitized paintings for visualization and correlation analysis of the achieved representations. In particular, we used variants of AlexNet \cite{10}, VGGNet \cite{11}, ResNet \cite{12}, which were originally developed for the task of object categorization for the ImageNet challenge \cite{13} and each of them raised the state of the art in that task when they were introduced. We adapted these networks for classifying 20 style classes. Our study included varying the training strategies (training from scratch on art data vs. using pre-trained models and fine-tuning them on art data), varying the network architecture, and data augmentation strategies.

\subsection*{Main Results}
\begin{figure*}[tbhp]
\centering
  \includegraphics[width=0.9\linewidth]{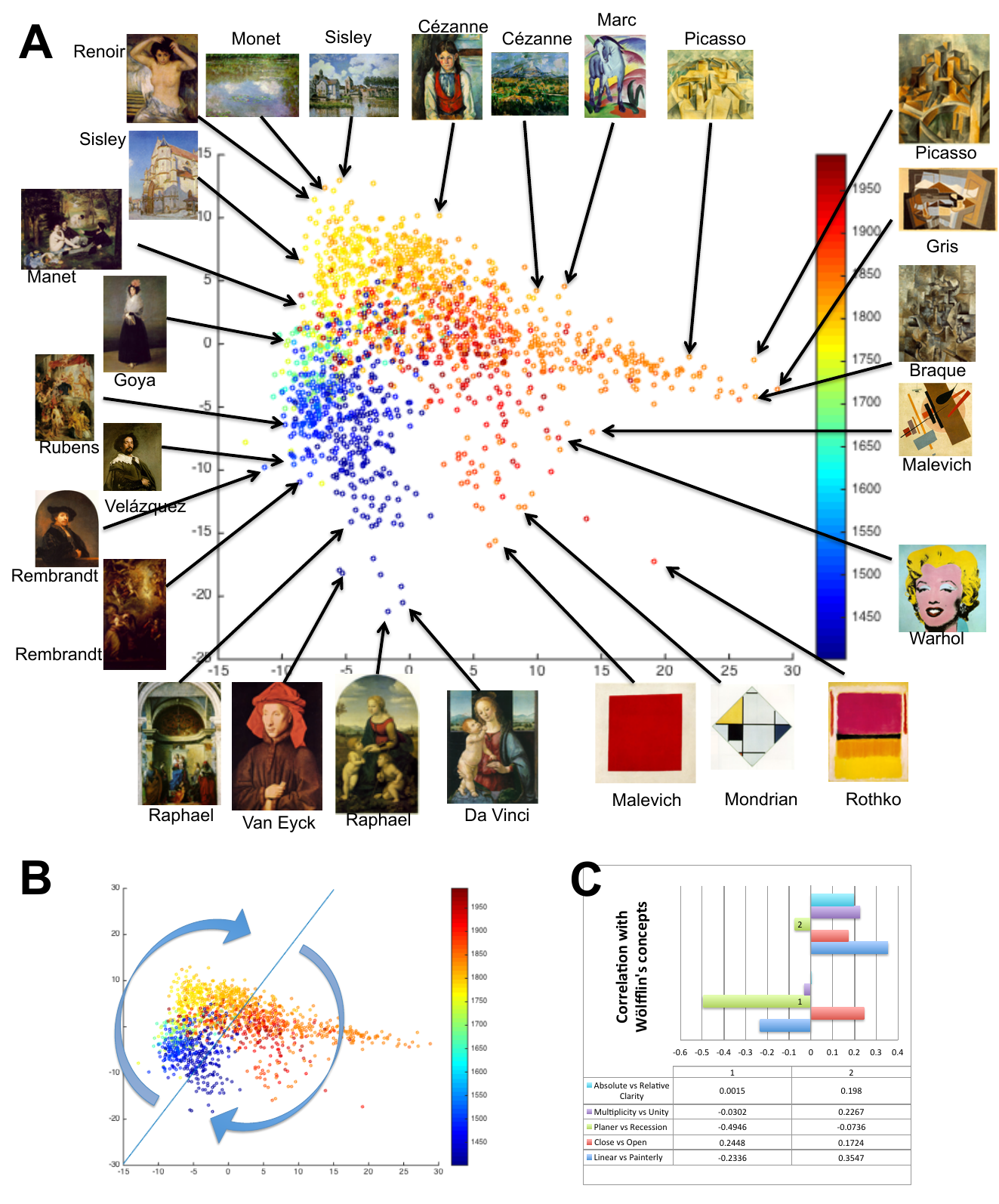}
  \caption{
  \footnotesize Modes of variations of the activation subspace showing smooth temporal transition and correlation with W\"olfflin's concepts. (A) First and second modes of variations of the AlexNet+2 model with paintings color-coded by date of creation. The first mode (the horizontal axis) seems to correlate with figurative art, which was dominant till Impressionism, vs. non-figurative, distorted figures, and abstract art that dominate 20th century styles. Another interpretation for this dimension is that it reflects W\"olfflin's concept of plane (to the right) vs.~recession (to the left). This axis correlates the most with W\"olfflin's concept of plane vs.~recession with -0.50 PCC. The vertical axis correlates with the linear (towards the bottom) vs. painterly (towards the top) concept (0.36 PCC). (B) The angular coordinate exhibits strong correlation with time (PCC of 0.69). (C) Correlation with W\"olfflin's concepts.}
  \label{F1}
\end{figure*}

Trained to predict styles, based only on noisy discrete style labels, without being given any notion of time, the machine encoded art history in a smooth chronology.  
The learned representation can be explained based on a handful of factors. The first two modes of variations are aligned with the concepts of linear vs.~painterly and planer vs.~recessional suggested by Heinrich W\"olfflin (1846-1945), and quantitatively explain most of the variance in art history, where temporal progression correlates radially across these modes (See Figure~\ref{F1}). The importance of these results is that they show that the selected art historian's theories about style change can be quantifiably verified using scientific methods.  The results also show that style, which appears to be a subjective issue, can be computationally modeled with objective means.


\section{Detailed Methodology}
\subsection{Challenges with Art Style Classification}

In contrast to the typical object classification in images, the problem of style classification in art has a different set of challenges that we highlight here. 
Style is not necessarily correlated with subject matter, which corresponds to existence of certain objects in the painting. Style is mainly related to the form and can be correlated with features at different levels, low, medium and high-level. As a result, it is not necessary that networks which perform better for extracting semantic concepts, such as object categories, would be perform as well in style classification. In the literature of object classification, deeper networks were shown to perform better \cite{11,12} since they facilitate richer representations to be learned at different levels of features. We do not know if a deeper network will necessarily be better in the style classification domain. This remains something to discover through our empirical study. However, the challenge, in the context of art style classification, is the lack of images on a scale of magnitude similar to ImageNet (million of images). The largest publicly available dataset, which we use, is only in the order of 80K images. This limitation is due to copyright issue, which is integral to the domain of art. Moreover, collecting annotation in this domain is hard since it requires expert annotators and typical crowd sourcing annotators are not qualified. 

Another fundamental difference is that styles do no lend themselves to discrete mutually exclusive classes as in supervised machine learning. Style transition over time is typically smooth, and style labels are after-the-fact concepts imposed by art historians, sometimes centuries later. Paintings can have elements that belongs to multiple styles, and therefore not necessarily identifiable with a unique style class. 

Paintings also come in wide variety of sizes. The canvas can span a couple of hundreds of square feet or can be as small as a couple of square inches. The aspect ratio can vary very significantly as well. A typical CNN requires its input to be resized to a fixed size retinal array. This bound to introduce geometric distortions, which can affect the composition, and loss of details related to surface texture carrying essential information about brush strokes. Both composition and brush strokes are essential to identify style. In this paper we do no address solving these issues. We mainly report the behavior of the studied model for style classification despite their limitation to address these issues.  

\subsection{Datasets}
Training-testing Set: We trained, validated, and tested the networks using paintings from the publicly available WikiArt dataset\footnote{Wikiart dataset http://www.wikiart.org}. This collection (as downloaded in 2015) has images of 81,449 paintings from 1,119 artists ranging from the fifteenth century to contemporary artists. Several prior studies on style classification used subsets of this dataset (e.g. \cite{4,5,6}). Originally WikiArt has 27 style classes. For the purpose of our study we reduced the number of classes to 20 classes by merging fine-grained style classes with small number of images, for example we merged cubism, analytical cubism, and synthetic cubism into one cubism class. Table~\ref{T:Styles} shows a list of the 20 style classes used and their mapping from the original WikiArt classes. We also ignored the symbolism style class since in the Wikiart collection this class contains various paintings from different styles that erroneously labeled as ``symbolism''.  In general, by visual inspection we notice that the style labels in the WikiArt collection is noisy and not accurate. However, it remains the largest publicly available collection. Other available collections, which might be cleaner in their labels, are typically much smaller in size for the purpose of training a deep network. We excluded from the collections images of sculptures and photography. The total number of images used for training, validation, and testing are 76,921 images. We split the data into training (85\%), validation (9.5\%) and test sets (5.5\%).

\begin{table*}[tbh]
\caption{Style classes used in learning the models}
\label{T:Styles}
\small
\center
\scalebox{0.8}{
\begin{tabular}{ |c|c|c|c|} 
 \hline
 \textbf{} & \textbf{Style Class}& \textbf{Number of Images}&\textbf{Merged styles} \\\hline
 1& Early Renaissance & 1391  &\\\hline
 2& High Renaissance  &  1343& \\\hline
 3& Mannerism and Late Renaissance  & 1279 & \\\hline
 4& Northern Renaissance & 2552 &\\ \hline
 5& Baroque & 4241 & \\ \hline
 6& Rococo & 2089& \\ \hline
 7& Romanticism & 7019 &\\ \hline
 8& Impressionism & 13060&\\ \hline
 9& Post-Impressionism & 6965 &Post Impressionism+Pointillism\\ \hline
 10& Realism & 11528 & Realism + Contemporary Realism + New Realism\\ \hline
 11& Art Nouveau  & 4334& \\ \hline
 12& Cubism & 2562& Cubism + Analytical Cubism + Synthetic Cubism \\ \hline
 13& Expressionism & 6736& \\ \hline
 14& Fauvism & 934 & \\ \hline
 15& Abstract-Expressionism & 2881 & Abstract Expressionism \& Action Painting\\ \hline
 16& Color field painting & 1615& \\ \hline
 17& Minimalism & 1337 &\\ \hline
 18& Na\"{i}ve art-Primitivism & 2405 &\\ \hline
 19& Ukiyo-e & 1167 &\\ \hline
 20& Pop-art & 1483 &\\ \hline
  &  Total number of Images&76921&\\
 \hline
\end{tabular}
}
\end{table*}

Visualization Set I: We used another smaller dataset containing 1485 images of paintings from the Artchive dataset\footnote{Artchive dataset http://www.artchive.com} to analyze and visualize the representation. Previous researches that have used this dataset (e.g. \cite{5}). We denote this dataset in this paper as the ``Visualization dataset I''. While the WikiArt collection is much bigger in size, this dataset contains a better representation of the important works of western art from 1400-2000AD by 60 artists. Therefore, we mainly use it to visualize and analyze the learned representation. We excluded from the collections images of sculptures and images containing partial details of paintings. We also collected art historian's rating annotations (scale of 1 to 5) for each of the W\"olfflin's pairs for 1000 paintings from this data set and use it in our correlation analysis. 

Visualization Set II: We also used 62K painting from the Wikiart dataset for visualization and analysis of the representations. We only included paintings that have date annotations for the purposes of visualization and temporal correlation. 
The two chosen visualization sets have complementary properties. While the Artchive visualization set represents knowledge about influential and important artists, the Wikiart is more arbitrarily chosen based on what is available in the public domain. The Artchive dataset lacks paintings from the 18th century and does not have broad sampling of early 19th paintings or post-war 20th century. In contrast, the Wikiart dataset is lacking certain works of art that art historians would consider important,  nevertheless it densely samples a broad range of styles. In addition, the Wikiart dataset has a large bias towards 20th century art and post WWII works. 

\subsection{Studied Deep Learning Models:}
We performed a comprehensive comparative study on several deep convolutional networks, adapted for the task of style classification. For the purpose of this paper, we report the results of three main networks: AlexNet~\cite{10}, VGGNet~\cite{11}, and ResNet~\cite{12}, as well as variants of them. All these models were originally developed for the task of object recognition for the ImageNet challenge~\cite{13} and each of them raised the state of the art when they were introduced. 

For the non-expert reader, here we briefly summarize the main features of each of these models. Deep convolutional networks in general consist of a sequence of layers of artificial neural units of different types. Convolutional layers apply learned filters (templates) to each location of its input image. These convolutional layers are interleaved with pooling layers, which aggregate the responses of the convolution layers. Typically, the sequence of convolution and pooling layers results in re-representing the visual information in the image as responses to a large number of learned filters applied to wider and wider regions of the image as the information propagates deeper in the network. Such filters are learned or tuned by the machine in response to the task in hand. Finally the responses are passed to a sequence of fully connected layers that acts as a classifier.

AlexNet~\cite{10} architecture consists of five consecutive convolution layers, interleaved with some pooling layers, followed by three fully connected layers, resulting in 1000 nodes representing the object classes in ImageNet. The convolutional filters have different sizes at each layer, starting from big filters (11x11) and reduced the filter size at following layers. In contrast VGGNet~\cite{11} adapted an architecture of fixed size filters (3x3) over a deeper sequence of convolutional layers. Residual Networks (ResNet~\cite{12}) introduced shortcut connection between the convolution layers outputs and later layers, which results in much deeper architectures, reaching over 150 layers.

In our study, in general for all the models, the final softmax layer, originally designed for the 1000 classes in ImageNet, was removed and replaced with a layer of 20 softmax nodes, one for each style class.  Two modes of training were evaluated, 1) training the models from scratch on the Wikiart data set, described above; 2) using a pre-trained model with ImageNet data and fine-tuned on the Wikiart data.  Fine-tuning is the standard practice when adapting well-performing pre-trained models to a different domain.

\subsection{Increasing the interpretability of the representation}

Having a large number of nodes at the fully connected layer allows the representation to project the data into a very high dimensional space where classification would be easy (especially in our case with only 20 classes), without forcing similar paintings across styles to come closer in the representation. To increase the interpretability of the representation, we force the network to achieve classification through a lower dimension representation. To achieve this, after training the network (whether from scratch or through fine-tuning), two more fully connected layers were added with a reduced number of nodes. These reduced dimensional layers force the representation to use a smaller number of degrees of freedom, which in turn forces paintings across styles to come closer in the representation based on their similarity. In particular, we added two layers with 1024 and 512 nodes to all the models, and the models were then fine-tuned to adjust the weights for the new layers. As will be shown later, adding these dimensionality reduction layers did not affect the classification accuracy. The experiments showed that gradually reducing the number of nodes in the fully connected layers forced the network to achieve a ``smoother'' and interpretable representation. It is important to emphasize that to achieve this effect without reducing the accuracy; the two new layers have to be added after the network is trained, then models are fine-tuned. Training the full architecture with the two extra layers, whether from scratch of fine-tuned, typically doesn't result in converging to similar accuracy.

We analyzed the modes of variations in the activation of each layer of the learned representation using Principle component analysis (PCA)~\cite{PCA}. We also analyzed the activation space using Independent Component Analysis (ICA)~\cite{ICA}. We analyzed the nonlinear manifold of activations through Laplacian Eigen Embedding (LLE)~\cite{LLE}. We chose these two techniques as widely used representatives of linear and nonlinear dimensionality reduction techniques, each providing a different insight about the learned representation. We also performed correlation analysis between the dimensions of the activation space and time as well as ground truth for W\"olfflin's concepts.

\section{Quantitative Comparative Results}

\begin{figure*}[tbh]
\centering
  \includegraphics[width=\linewidth]{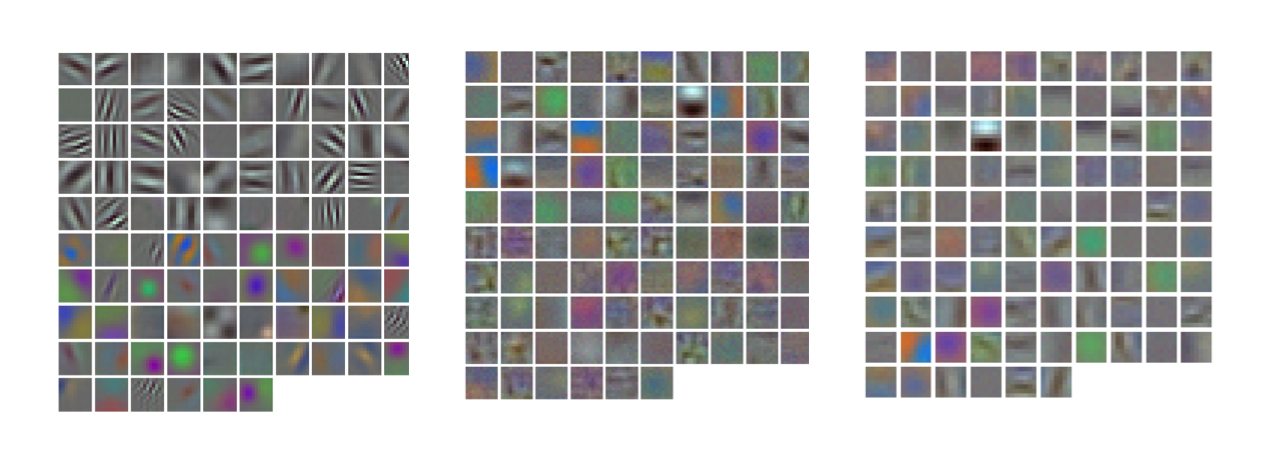}
  \caption{Comparison of the filters learned on object classification vs style classification. Left: Typical filters of AlexNet trained on ImageNet for object classification. Center and Right: Filters of AlexNet trained for style classification using 11x11 and 7x7 dim filters.}
  \label{FS1}
\end{figure*}

Table~\ref{T:Compare} shows the classification accuracy of different models using both pre-training with fine-tuning, and training from scratch. In all cases, the pre-trained and fine-tuned networks achieved significantly better results than their counterparts that are trained from scratch (7\% to 18\% increase). This is not surprising and consistent with several models that adapted and fine-tuned pre-trained networks for different domains. However, learned filters when the network trained from scratch on the art domain were significantly different from the ones typically learned on ImageNet, which typically shows Gabor-like and blob-like filters. Figure~\ref{FS1}  shows visualization of the filters of AlexNet when trained on ImageNet compared to same filters trained on WikiArt for style classification. While it is hard to interpret the filters trained for style classification, we do not observe oriented-edge-like filters except for a horizontal edge filter.  This emphasizes the difference in nature between the problems and suggests that the better performance of the fine-tuned models could be out-performed if sufficient data is available to train a style-classification network from scratch on art data only.

\begin{table}[tbh]
\caption{Comparison of classification of different models and different training methodologies}
\label{T:Compare}
\small
\scalebox{0.95}{
\begin{tabular}{ |l|l|l|l| } 
 \hline
 \textbf{Network} & \textbf{Architecture}& \textbf{Trained from scratch}&\textbf{{Pre-trained \& Fine-tuned}} \\\hline
 AlexNet& Original:5 Conv layers+3 FC layers & 48.1\%  &58.2\%\\ \hline
 AlexNet+2& Adding 2 reduced FC layers  &  47.2\%& 58.3\%\\ \hline
 VGGNet& 13 Conv layers + 3 FC layers  & 51.6 \% & 60.1\%\\\hline 
 VGGNet+2& Adding 2 reduced FC layers& 55.2\% &62.8\%\\ \hline
 ResNet& 152 Conv layers & & 63.7\%\\ \hline
 & 50 Conv layers & 45.0\%& \\ \hline
 ResNet+2& 152 Conv layers +2 reduced FC layers &  &60.2\%\\ \hline
 & 50 Conv layers + 2 reduced FC layers & 48.0\%&\\ \hline
\end{tabular}
}
\end{table}

Increasing the depth of the network only added no more than 5\% to the accuracy from AlexNet with 5 convolutional layers to ResNet with 152 convolutional layers. In the case for learning from scratch, increasing the depth did not improve the results where a ResNet with 50 layers performed worse than an AlexNet with only 5 convolution layers. VGGNet with 13 convolutional layers performed only 2-3\% better than AlexNet. Increasing the depth of VGGNet did not improve the results. This limited gain in performance with increase in depth, in conjunction with the difference in the learned filters, suggests that a shallow network might be sufficient for style classification along with better filter design.

\subsubsection*{Effect of adding layers with reduced dimensionality:}
The experiments showed that adding extra fully connected layers while gradually reducing the number of nodes in them forces the networks to achieve a ``smoother'' and more interpretable representation. Having a large number of nodes at the fully connected layers allows the representation to project the data into a very high dimensional space where classification would be easy (specially in our case with only 20 classes), without necessarily enforcing similar paintings across styles to come closer in the representation based on their similarity. We quantified this phenomenon by examining the dimensionality of the subspace of the activation (using Principle Component Analysis (PCA~\cite{PCA}) of the visualization dataset using two measures: 
1) The number of components needed to preserve 95\% of the variance.  
2) The variance retained with the first two PCA dimensions. 
We also evaluated the accuracy of the expanded models to see if the added reduced layers resulted in any loss of accuracy. In most of the cased the added layers enhanced the accuracy (see Table~\ref{T:Compare})

\begin{table}[tbh]
\caption{The effect of adding two reduced dimension layers on the representation (subspace
dimensionality and variance)}
\label{T:SI3}
\small
\scalebox{0.7}{
\begin{tabular}{|l|l|c|c|c|c|c|c|}\hline
{Model} & {Training Strategy} & \multicolumn{3}{c|}{Original Architecture} & \multicolumn{3}{c|}{Adding two dimensionality reduction layers} \\\cline{3-8} & &number of nodes &subspace dim $^1$ &retained variance $^2$ & number of nodes &subspace dim$^1$ &retained variance$^2$ \\\hline
{AlexNet}& Pre-trained & {4096} & {201} & {21.71} & {512}& {9}& {59.64}\\
& \& Finetuned &&&&&& \\\hline
AlexNet&From Scratch&4096&397&35.62&512&10&62.78\\\hline
{VGGNet}& Pre-trained &{4096} &{55} &{49.52} &{512}&{7}&{66.87}\\
& \& Finetuned &&&&&& \\\hline
VGGNet&From Scratch&4096&36&51.16&512&7&72.52\\\hline
{ResNet}& Pre-trained &{2048}$^3$ &{491} &{17.53} &{512}&{6}&{73.71}\\
& \& Finetuned &&&&&& \\
\hline
\multicolumn{8}{|l|}{$^1$ Subspace dim: Number of principle components cumulatively retaining 95\% of variance.} \\
\multicolumn{8}{|l|}{$^2$ Retained variance: Percentage of variance retained by the first two principle components. } \\
\multicolumn{8}{|l|}{$^3$ ResNet does not have FC layers. This is the number of the nodes in the last pooling layer. } \\ \hline
\end{tabular}
}
\end{table}

\begin{figure*}[tbh]
\centering
  \includegraphics[width=\linewidth]{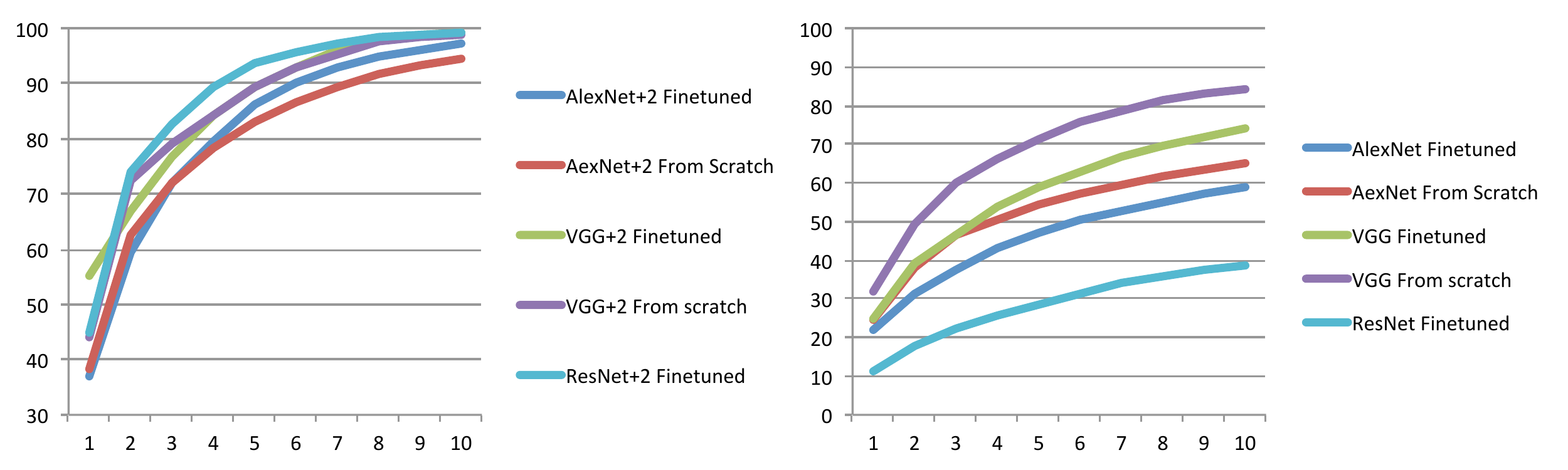}
  \caption{Cumulative retained variance by the first 10 PCA dimensions of the activation space at last FC layer (512 nodes). Left: models after adding the dimensionality reductions layers. Right: the original models.}
  \label{FS2}
\end{figure*}

 Table~\ref{T:SI3} shows that adding two reduced-dimension layers effectively and consistently reduced the dimensionality of the subspace of the data while preserving the classification  accuracy. The reduction is significant for AlexNet where 9 or 10 dimensions retained 95\% of the variance compared to 201 and 397 dimensions for the cases of fine-tuned and learned from scratch networks respectively, with around 60\% of the variance retained in the first two dimension.  Interestingly, the representation achieved by VGGNets already has reduced dimension subspaces compared to the AlexNet and ResNet. However, adding the reduced dimension layers for VGG significantly lowered the subspace dimension (only 7 dimensions retain 95\% of the variance) while improving its classification accuracy between 2-4\%. The maximum reduction in subspace dimensionality was in ResNet where the dimension of the subspace retaining 95\% of the variance was reduced from 491 to only 6 with 74\% of the variance in the first two dimensions. Figure~\ref{FS2} shows the cumulative retained variance for the first 10 PCA dimensions of the activation subspace for all the models before and after adding the two reduced dimensions layers.

\section{Shape of Art History: Interpretation of the Representation}

This section focuses on analyzing, visualizing, and interpreting the activation space induced by the different networks after trained to classify style. 

We define the activation space of a given fully connected layer as the output of that layer prior to the rectified linear functions.  In particular, in this paper, we show the analysis of activation of the last reduced dimension fully connected layer prior to the final classification layer, which consists of 512 nodes. We use the activations before the rectified linear functions in all the networks.

\subsubsection*{Few Factors Explain the Characteristics of Styles:}
The learned representation by the machine shows that there are a few underlying factors that can explain the characteristics of different styles in art history. Using Principle Component Analysis, we find that only fewer than 10 modes of variations can explain over 95\% of the variance in the visualization set in all of the studied models with additional reduced fully connected layers. In ResNet and VGGNet the number of these modes is as low as 6 and 7 respectively. In all of the networks, the first two modes of variations explained from 60\% to 74\% from the variance in the various models in visualization set I (Table~\ref{T:SI3}). Moreover, it is clear from the visualizations of both the linear and nonlinear embedding of the activation manifold in various models that art dated prior to 1900 lie on a plane (subspace of dimension 2).  

Consistent results are achieved by analyzing the 62K painting from the Wikiart data set where it was found that subspaces of dimensions 10, 9 and 7 retain 95\% of the variance of the activation for AlexNet+2, VGGNet+2, ResNet+2 respectively (Table~\ref{T:SI4}). The consistency of results in all the studied networks and the two datasets (varying in size from $\approx$ 1500 to 62K paintings) imply that the existence of a small number of the underlying factors explaining the representation is an intrinsic property of art history, and not just an artifact of the particular dataset or model. 

We will start our interpretation of these dimensions by investigating the time correlation with these dimensions, followed by correlation analysis with W\"olfflin's concepts.

\begin{table*}[htb]
\caption{Temporal correlation with the first two PCA dimensions and the first LLE dimensions of
the activation space in different models}
\label{T:TempCor}
\small
\center
\begin{tabular}{|l|l|r|r|r|r|}\hline
&&\multicolumn{4}{c}{Pearson correlation coefficient with time} \\\hline
model&training&$1^{st}$ PCA dim &$2^{nd}$PCA dim & Radial & $1^{st}$LLE dim\\\hline
AlexNet+2&Fine-tuned&0.4554&0.5545&0.6944&0.7101\\
&From scratch&-0.5797&0.2829&0.6697&0.6723\\
VGGNet+2&Fine-tuned&-0.2462&0.5316&0.5659&-0.4012\\
&From scratch&0.5781&0.3165&0.6239&-0.6532\\
ResNet+2&Fine-tuned&-0.6559&0.4579&0.7712&0.8130\\\hline
\end{tabular}
\end{table*}

\subsubsection*{Smooth Temporal Evolution of Style:}
\begin{figure*}[tbhp]
\centering
  \includegraphics[width=\linewidth]{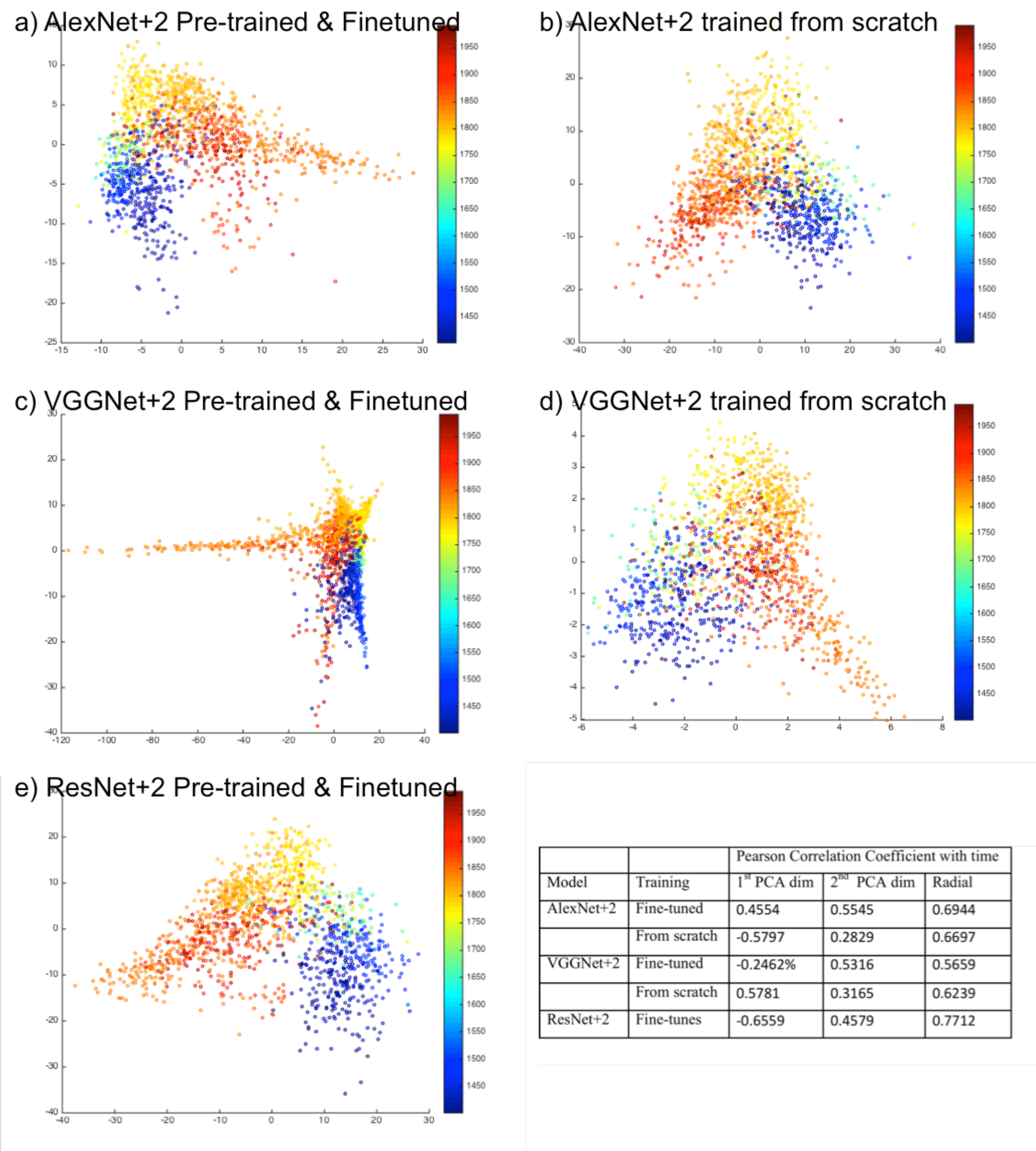}
  \caption{First two modes of variations of different learned representations and their temporal correlations. Every point is a painting, color-coded by the year of creation. Dataset: VS-I.}
  \label{FS3}
\end{figure*}

All the learned representations show smooth transition over time. Figure~\ref{FS3} shows the paintings in the visualization set I, projected into the first two modes of variations of the activations of the last reduced dimension fully connected layer for different networks. The paintings are color-coded based on their date of creation. The plots consistently show that the networks learned smooth temporal representations that reflect the historical progress in style.  This is despite the fact that the networks are trained only with images and their discrete style labels. No information was provided about when each painting was created, when each style took place, which artist created which painting, nor how styles are related (such as style x is similar to style y, or came after or before style z). Despite the lack of all this information, the learned representations are clearly temporally smooth and reflect high level of correlation with time. 

Visualizing these modes of variations reveals very interesting ways of how the networks arrange the data in a temporal way. For example, for the case of AlexNet+2 model, as shown in Figure~\ref{F1}-A, the data is arranged in the plot in a radial clock-wise way around the center to make a complete circle in this 2D projection starting with Renaissance and ending with Abstract Art. We can see the progress following the plot in a clock-wise way from Italian and Northern Renaissance at the bottom, to Baroque, to Neo-classicism, Romanticism, reaching to Impressionism at the top followed by Post impressionism, Expressionism and Cubism. The cycle completes with Abstract and Pop Art. The explanation of how the network arranged the representation this way will become clearer next when we show the results of nonlinear activation manifold analysis.

We quantified the correlation with time in different ways. We measure the Pearson Correlation Coefficients (PCC) with each of the principle components, as well as with each dimension of the nonlinear manifold embedding.  Table~\ref{T:TempCor} shows the temporal correlation using PCC with the first two PCA dimensions and the first LLE embedding dimension for different models. We also computed the correlation of the radial coordinate -- of each painting in the projection to the first two principle components-- with time. The temporal correlation obtained in the representation is consistent among the different networks. 

Taking a fine-tuned AlexNet+2 as example, the first two dimensions have very strong correlation with time, 0.46 and 0.55 Pearson correlation coefficients respectively. A linear combination of these two dimensions with convex weights proportional to their Pearson correlation coefficients results in a single dimension that has 0.71 correlations with time (Figure~\ref{F1}-B).  This cycle around the plot from Renaissance till abstraction and Pop art suggests that the angular coordinate around the plot would reflect correlation with time. In fact the angular coordinates have a PCC of 0.69 with time.  The strongest temporal correlation is in the case of ResNet with 0.76 PCC radial correlations.  This conclusion is consistent among all the networks that we tested. This conclusion shows that style changes smoothly over time and proves that noisy discrete style labels are enough for the networks to recover temporal arrangement, due mainly to visual similarities as encoded through the learned representation in a continuous activation space. 

Strong temporal correlation also exists with the modes of variations in Visualization set-II. However, in this case, it is not necessarily the first two modes of variation that are the ones with the highest temporal correlation. This is due to the bias in this dataset towards 20th century work. Table~\ref{T:SI4} shows the PCC with time for each of the dimensions for three fine tuned networks. 

\begin{table}[tbh]
\caption{Analysis of the modes of variations of the activations on 62K paintings from the Wikiart collection in three models and their correlation with time (Pearson Correlation Coefficients) .
The top two time-correlated modes are highlighted in bold.}
\label{T:SI4}
\scalebox{0.77}{
\small
\begin{tabular}{ |l|c|c||c|c|c|c|c|c|c|c|c|c| }\hline
 & Subspace Dim $^1$ & Retained Variance $^2$  & \multicolumn{10}{|c|}{Correlation with time } \\
 &&& 1 & 2 & 3 & 4 & 5 & 6 & 7 & 8 & 9 & 10 \\ \hline
 AlexNet+2 & 10 & 51.13 & 0.12 & {\bf 0.66} & 0.24 & {\bf -0.43} & -0.23 & 0.04 & 0.08 & -0.05 & -0.03 & 0.00 \\\hline
 VGGNet+2 & 9 & 43 & -0.15 & -0.14 & {\bf -0.46} & -0.07 & {\bf -0.59} & 0.28 & 0.02 & 0.07 & 0.02 &  \\\hline
 ResNet+2 & 7 & 66.80 & -0.03 & {\bf -0.72} & 0.20 & 0.22 & {\bf 0.30} & -0.01 & 0.06 &  &  &  \\\hline
 \multicolumn{13}{|l|}{$^1$ Subspace Dim: Number of principle components retaining 95\% of the variance} \\
 \multicolumn{13}{|l|}{$^2$ Retained Variance: Variance retained by first two dimensions }\\ \hline
\end{tabular}
}
\end{table}

\subsubsection*{Interpretation of the Modes of Variations - Relation to W\"olfflin's Pairs: }
Visualizing the modes of variations reveals very interesting ways of how the networks were able to consistently capture evolution and characteristics of styles.
Here we take the fine-tuned AlexNet+2 as example, however similar results can be noticed in other models. The first mode (the horizontal axis in Figure~\ref{F1}) seems to correlate with figurative art, which was dominant till Impressionism, vs. non-figure, distorted figures, and abstract art that dominates 20th century styles. Another interpretation for this dimension is that it reflects W\"olfflin's concept of planar (to the right) vs. recession (to the left). This axis correlates the most with W\"olfflin's concept of planar vs recession with -0.50 Pearson correlation coefficient. To a lesser degree, this horizontal axis correlates with closed vs. open (0.24 PCC) and linear vs. painterly (-0.23 PCC). This quantitative correlation can be clearly noticed by looking at the sampled paintings shown in Figure~\ref{F1}, where we can see that the horizontal axis is characterized by planar, open, and linear form to the right (as in Cubism, and Abstract) vs. recession, closed, and painterly to the left (as appears in some of the Baroque paintings at the extreme left) 

\begin{table*}[tbhp]
\caption{Correlation with W\"olfflin's concepts. Pearson Correlation Coefficient of the first two PC
dimensions and the first two LLE dimensions of the activation space and W\"olfflin's
concepts. The concepts with maximum correlation with each dimension are shown.}
\label{T:WCor}
\small
\begin{tabular}{clrrrr}\hline
&&\multicolumn{4}{c}{Pearson correlation coefficient(absolute values) with W\"olfflin's
concepts}\\\hline
{Model}&{Training}&$1^{st}$ PCA dim  &$2^{nd}$PCA dim &  $1^{st}$LLE dim & $2^{nd}$LLE dim\\
&&Planar vs.&Linear vs.& Planar vs.&and Linear vs.\\
&&Recession &Painterly & Recession & Painterly\\\hline
AlexNet+2&Fine-tuned&0.4946&0.3579&-0.501&0.3216\\
&From scratch&0.5129&0.3272&0.4930&-0.3111\\
VGGNet+2&Fine-tuned&0.3662&0.2638&0.4512&0.2646\\
&From scratch&0.4621&0.4000&0.4897&0.3174\\
ResNet+2&Fine-tuned&0.5314&0.4795&0.5251&0.4158\\\hline
\end{tabular}
\end{table*}

The second mode of variations (the vertical axis) has 0.36 Pearson correlation coefficient with the linear (towards the bottom) vs. painterly (towards the bottom) concept. We can clearly see the smooth transition from linear form in Renaissance at the bottom towards more painterly form in Baroque to the extreme case of painterly at Impressionism at the top. Then we can see the transition back to linear form in abstract and Pop art styles. Projecting the data into these two dominant modes of variations, which are aligned with plane vs. recession and linear vs. painterly, gives an explanation to why this representation correlates with time in a radial fashion.

The correlation between the first two modes of variations and the concepts of planar vs. recession and linear vs. painterly is consistent among all the representations learned by all tested networks, whether pre-trained or trained from scratch. In all cases the first mode of variations correlates the most with the concept of planar vs. recession while the second mode correlates the most with linear vs. painterly. Table~\ref{T:WCor} shows the Pearson correlation coefficients with these concepts. Full correlation results with all dimensions are shown in the appendix.

\begin{figure*}[tbhp]
\centering
  \includegraphics[width=\linewidth]{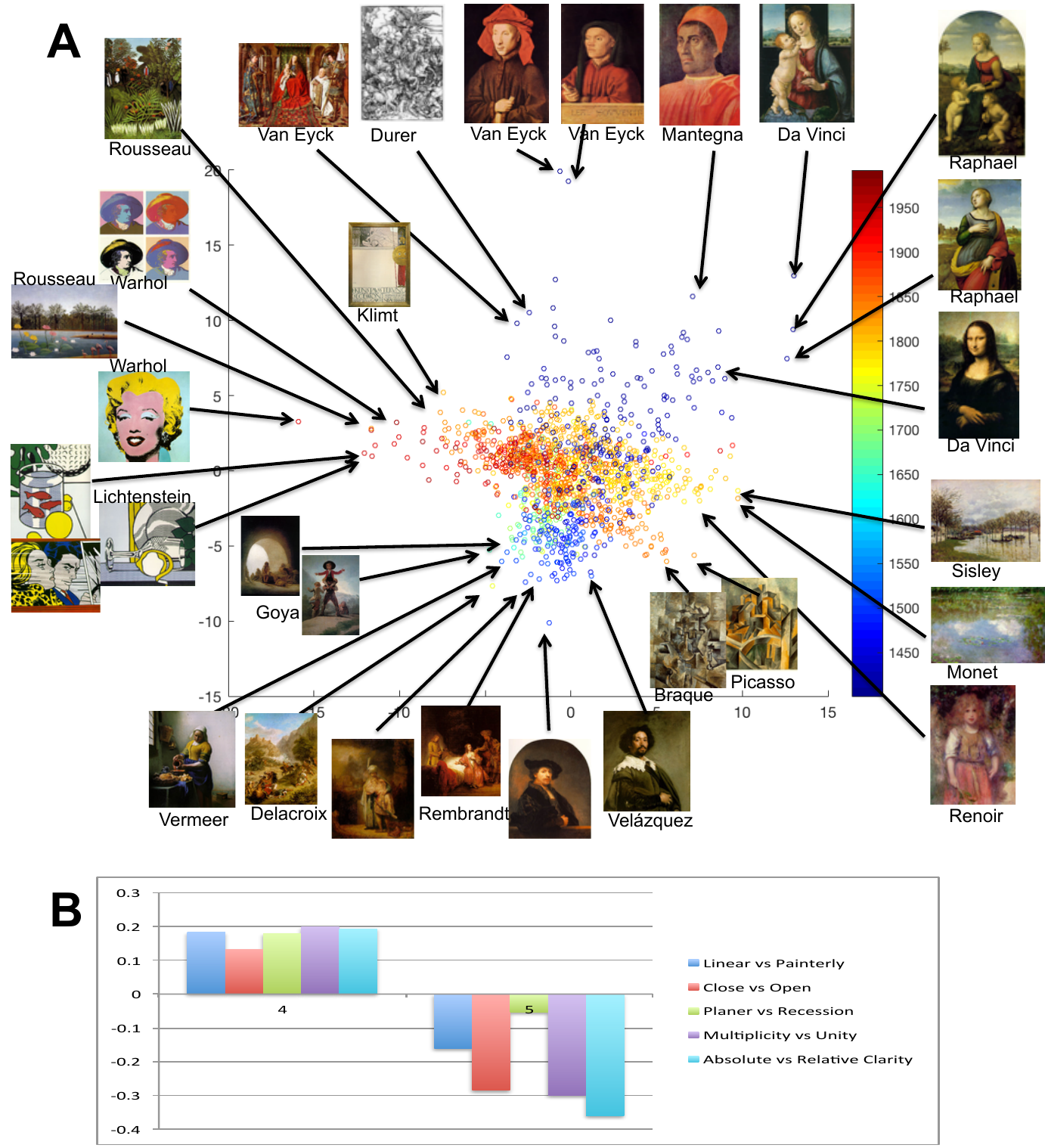}
  \caption{Separation between Renaissance and Baroque: (A) The fourth and fifth modes of variations show separation between Renaissance and Baroque. (B) Correlation between these two modes of variations and W\"olfflin's concepts, which confirms with  W\"olfflin's hypothesis.}
  \label{F5}
\end{figure*}

The fourth and fifth dimensions of AlexNet+2 representation spread away strongly the Renaissance vs Baroque styles and put other styles in perspective to them (Figure~\ref{F5}).  The fifth dimension in particular (and the fourth dimension to a lesser degree) correlates with relative vs. absolute clarity, unity vs. multiplicity, open vs. closed, and painterly vs. linear form from top to bottom (with PCC 0.36, 0.30, 0.28, 0.16 respectively). This is consistent with W\"olfflin's theory since he suggested exactly the same contrast between these concepts to highlight the difference between Renaissance and Baroque.  In Figure~\ref{F5}-A , the Renaissance style appears at the top (absolute clarity, multiplicity, closed, linear form) while the Baroque appears at the bottom (relative clarity, unity, and open, painterly form). We can see in the figure that Impressionism and Cubism are at the top half of the plot since they share many of these same concepts with Baroque. The fourth dimension seems to separate Impressionism and Cubism to the right from abstraction and Pop art to the left. 

Looking at the 4th, 5th, and 6th dimensions of the embedded activation manifold, we can notice two orthogonal planes characterizing art prior to 1900 (Figure~\ref{FS6}). One plane spans Renaissance vs. Baroque while an orthogonal plane spans Impressionism vs. Post-Impressionism.

\begin{figure*}[tbhp]
\centering
  \includegraphics[width=\linewidth]{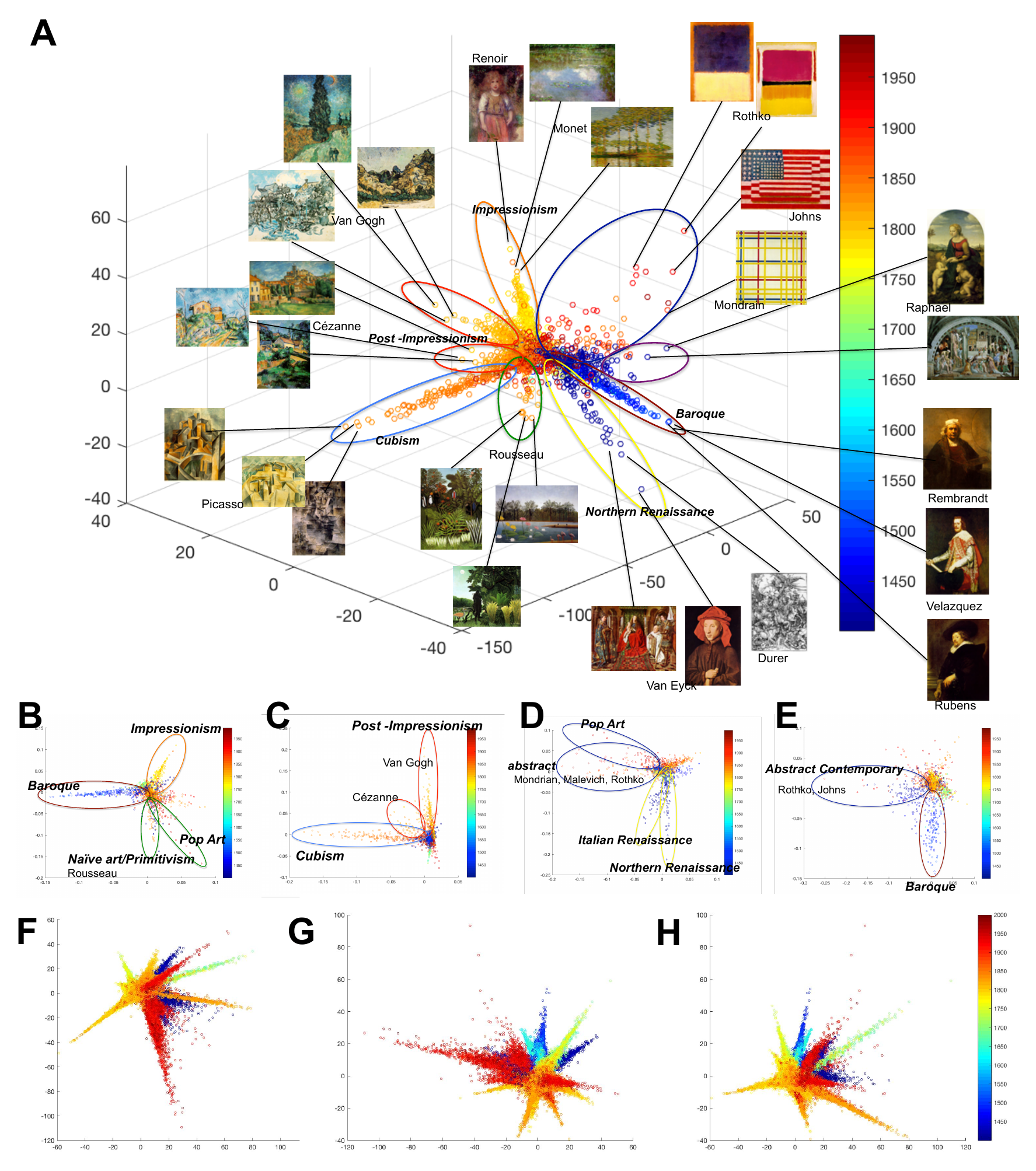}
\caption{\small The representation could discover distinctive artists for each style. (A)The first three modes of variations of the VGGNet activations. Distinctive artists representing each style are identified by the representation and pulled away from the cloud at the center.  (B-E) Factorization of the activation space using Independent Component Analysis into 7 maximally independent axes, which show alignment with styles (more details in Figure~\ref{FICA}). (F-H) The top three modes of variations in the VGG network activation of 62K works of art from the Wikiart collection (projected pairwise as dimensions 1-2, 2-3, 3-1 from left to right).}
\label{FVGG}
\end{figure*}

\subsubsection*{Discovering Limitations of W\"olfflin's Concepts:}

\begin{figure*}[tbhp]
\centering
  \includegraphics[width=\linewidth]{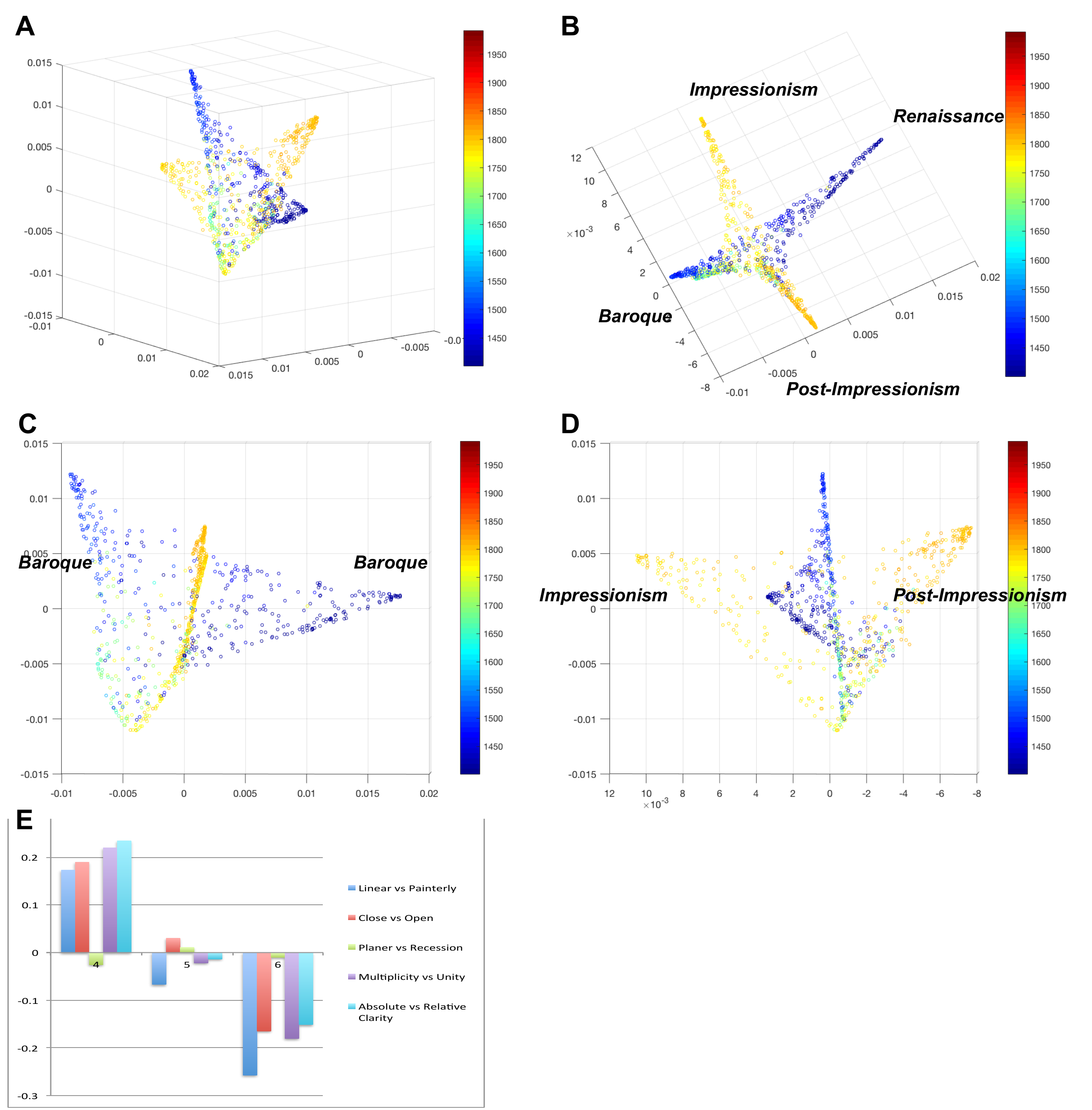}
  \caption{Two orthogonal planes characterize art prior to 1900 in the learned Representation. Renaissance-Baroque plane seems orthogonal to the Impressionism-Post-Impressionism plane in the embedding of the activation manifold of fine-tuned AlexNet+2 model is shown. (A,B) Two different view of the  4th, 5th, 6th dimensions are shown. (C) the 4th and 6th dimensions span the Renaissance-Baroque differences. (D) 5th and 6th dimensions. The 5th dimension spans the Impressionism-Post-Impressionism differences.  Interestingly, the  5th dimension has very small correlation with  W\"olfflin's concepts. (E) Correlation with  W\"olfflin's concepts}
  \label{FS6}
\end{figure*}

  Interestingly, not all modes of variations explaining the data correlate with W\"olfflin's concepts. In all learned representations, one of the first five modes of variation always has close-to-zero linear correlation with all W\"olfflin's concepts. A notable example is the fifth dimension of the embedded activation manifolds, which separates Impressionism from Post-Impressionism, and has almost zero linear correlation with W\"olfflin's pairs (Figure~\ref{FS6}). This implies that the separation between these styles is not interpretable in terms of W\"olfflin's pairs.

\subsubsection*{Identification of Representative Artists:}
Visualizing the different representations shows that certain artists were consistently picked by the machine as the distinctive representatives of their styles, as they were the extreme points along the dimensions aligned with each style. This is visible in the first three modes of variations of the representation learned by the VGGNet, shown in Figure\ref{FVGG}-A, which retains 77\% of the variance . Besides the observable temporal progress, the representation separates certain styles and certain artists within each style distinctively from the cloud at the center as non-orthogonal axes.

We can see the Northern Renaissance in the yellow ellipse with the majority of the paintings sticking out being by Van Eyck and D\"urer. The Baroque in the black ellipse is represented by Rubens, Rembrandt, and Velázquez. The orange ellipse is Impressionism and at its base are Pissarro, Caillebotte, and Manet as the least painterly of the type, ending with Monet and Renoir as most painterly on the end of the spike. The two red circles are Post-Impressionism, and in particular one is dominated by Van Gogh, and the other by C\'ezanne who forms the base for the spike of Cubism in the light blue ellipse. This spike is dominated by Picasso, Braque, and Gris; and goes out to the most abstract Cubist works. Most interestingly the representation separates Rousseau, as marked in the green ellipse, which is mainly dominated by his work. Consistent results can be seen when plotting the embedding of the activation of the whole 62K paintings in the Wikiart dataset collection as can be seen in Figure~\ref{FVGG} F-H.


Applying Independent Component Analysis (ICA)~\cite{ICA,FASTICA} to the activation space results in finding non-maximally independent non-Gaussian orthogonal axes\footnote{Independent Component Analysis (ICA) transforms activation to a new embedding space in which transformed data become statistically independent as much as possible and the ICA achieve this by finding non-orthogonal basis in original activation space maximizing independence criteria.}. Figure~\ref{FVGG} B-E shows the ICA factorization of the VGG network activation space. Here, the same subspace that retains 95\% of the variance is factorized by ICA. This results in 7 maximally independent components shown in Figure~\ref{FVGG}-E. 

\begin{figure*}[tbh]
\centering
  \includegraphics[width=\linewidth]{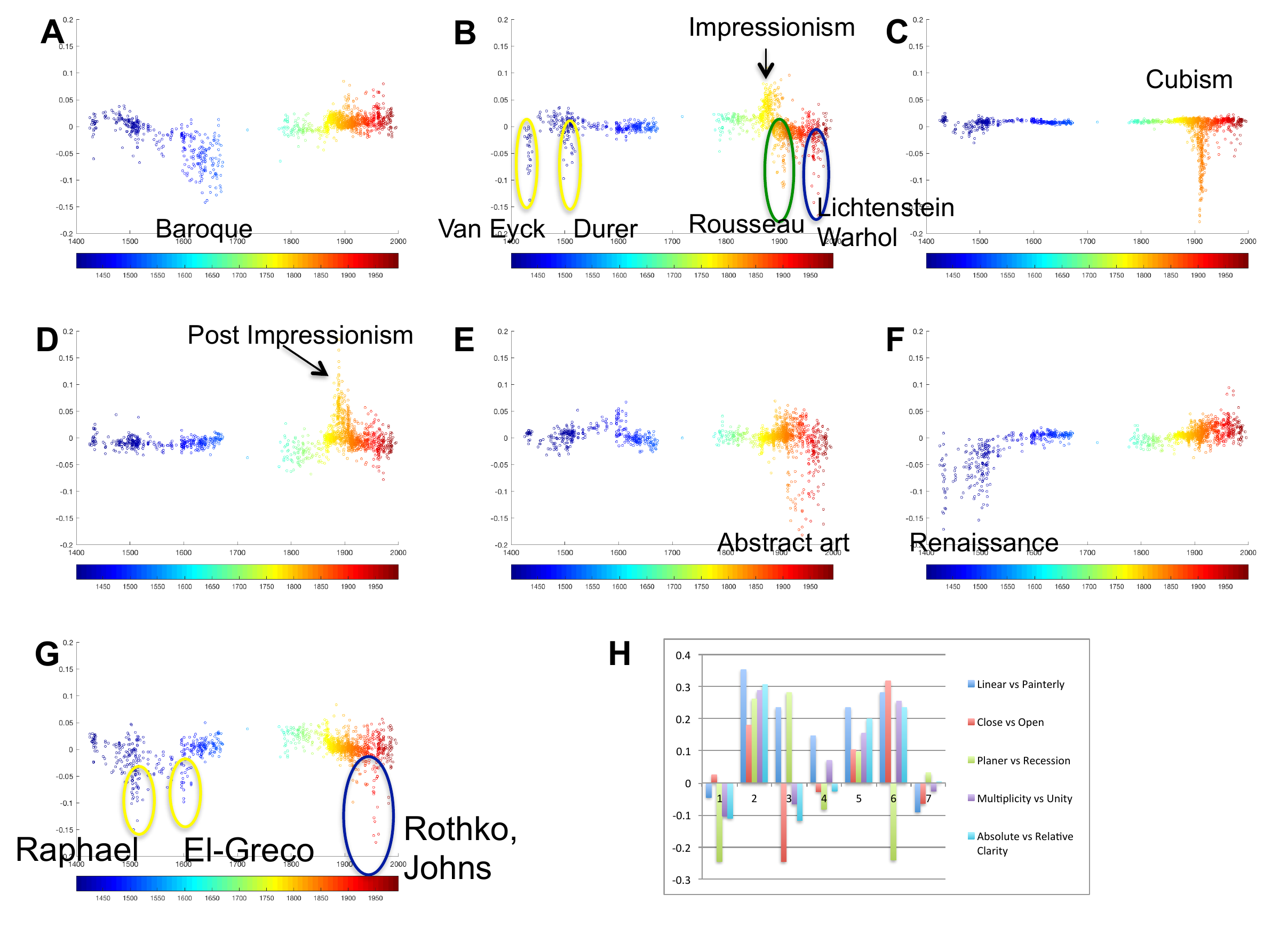}
  \caption{Independent component factorization of the activation of VGG network representation achieved by ICA. A-G: Each independent component vs time (1400-2000AD). H: Correlation (PCC) with W\"olfflin's concepts. Some components are aligned with specific styles (A: Baroque, C: Cubism, D: Post-Impressionism, E: abstract art, F: Renaissance) and other components contrast several styles (B and G). }
  \label{FICA}
\end{figure*}

We also show each of these components against the time axis in Figure~\ref{FICA}. Some components are aligned with specific styles (A: Baroque, C: Cubism, D: Post-Impressionism, E: abstract art, F: Renaissance) and other components contrast several styles (B and F). This factorization facilitates quantitatively discovering the artists who represent each style, as their works appear as the extreme points on the axes of each style. It also highlights which styles are opposite to each other along each of these axes \footnote{Since the ICA algorithm (Fast ICA~\cite{FASTICA}) used in this experiment has a limitation to capture over-completeness, which implies existence greater number of fat tails than dimension of the space, our experiment found only some of axes are aligned with styles. However, still the results show that ICA has merits to find some of interesting art historical facts and provide quantitatively measurement of them, such as what art works or which artists are most representative within each of styles.}.  

The fact that the representation highlights a certain representative artist or artists for each style -- among hundreds or thousands of paintings by many artists in each style  -- emphasizes quantitatively the significance of these particular artists in defining the styles they belong to.


\subsubsection*{Interpretations of the Activation Manifold and Understanding Influences:}
Using nonlinear dimensionality reduction of the activation allows us to achieve an embedding of the activation manifold of the visualization data sets, which reveals interesting observations as well. In particular we used Local Linear Embedding (LLE)~\cite{LLE}. LLE, as other similar techniques, is based on building a graph of data points, where each point is connected to its nearest neighbors, and using such graph to embed the data to a low dimensional space where it can be visualized. In our case the data points are the activation of a given painting as it goes through the deep network, and measured at the last fully connected layer, and this results in an embedding of the activation manifold. Controlling the number of neighbors of each painting in constructing the graph allows for controlling connections between paintings. If this parameter is set to a small number, this results in accentuating distinctive paintings and trends. If we set this parameter to a large number, we allow more paintings to connect and we can capture the overall trend.

\begin{figure*}[tbph]
\centering
  \includegraphics[width=\linewidth]{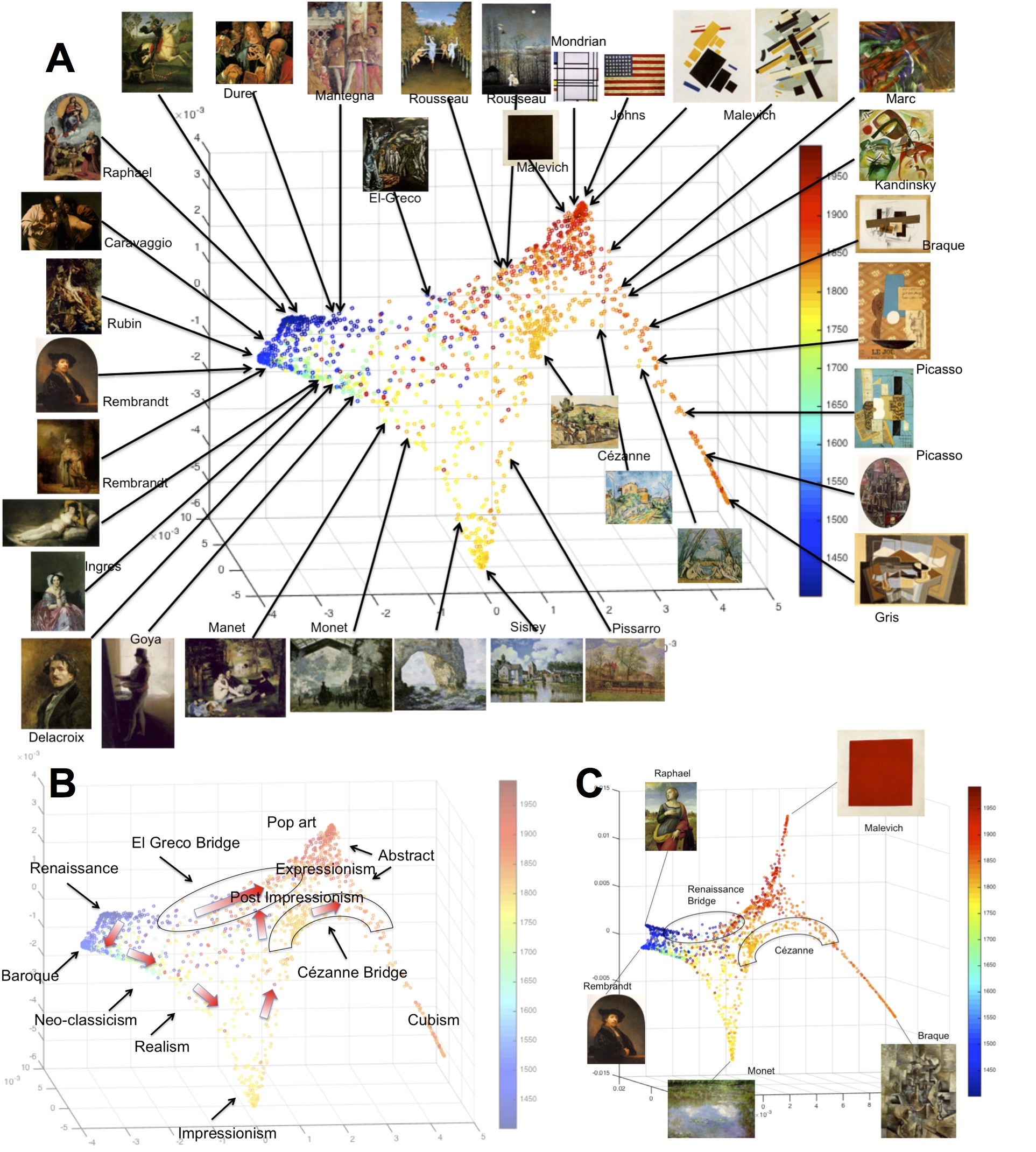}
\caption{Interesting connections discovered in the activation manifold. (A) Example of activation manifold of AlexNet+2.  Paintings color coded by date on the surface of the activation manifold showing smooth transition over time. (B) Transitions between art movements and important connections. (C) Accentuated version of the manifold highlighting five major styles: Renaissance, Baroque, Impressionism, Cubism, and abstract. }
\label{FLLE}
\end{figure*}

The first dimension of the LLE representation of all the models exhibits strong temporal correlation, as shown in Table~\ref{T:TempCor}. For example, the first LLE dimension in AlexNet+2 representation has 0.71 PCC with time that of ResNet+2 has 0.81 PCC with time. Also, similar to the modes of variations, the first two LLE dimensions shows strong correlation with W\"olfflin's concepts planar vs. recession and linear vs. painterly respectively, see Table~\ref{T:WCor}.
Looking at the representation learned by AlexNet+2, as an example, we can notice that on the surface of the embedded activation manifold, different styles are arranged in the learned representation in a way that reflects their temporal relation as well as stylistic similarity (see Figure~\ref{FLLE}-A)\footnote{The representation is achieved by setting the neighborhood size to 100 to capture the overall trend.}. The temporal relation between styles is not just linear historical order, since elements of some styles can appear later in other styles. For example we can see linear temporal progress from Renaissance, to Baroque, to Romanticism, to Realism to Impressionism. However we can also see direct connection between Renaissance to Post-Impressionism and 20th century styles. This can be explained since the use of linear (vs. painterly) and planar (vs. recessional) concepts in these later styles echo these elements from Renaissance style(Fig~\ref{FLLE}-B). 

\begin{figure*}[tbph]
\centering
  \includegraphics[width=\linewidth]{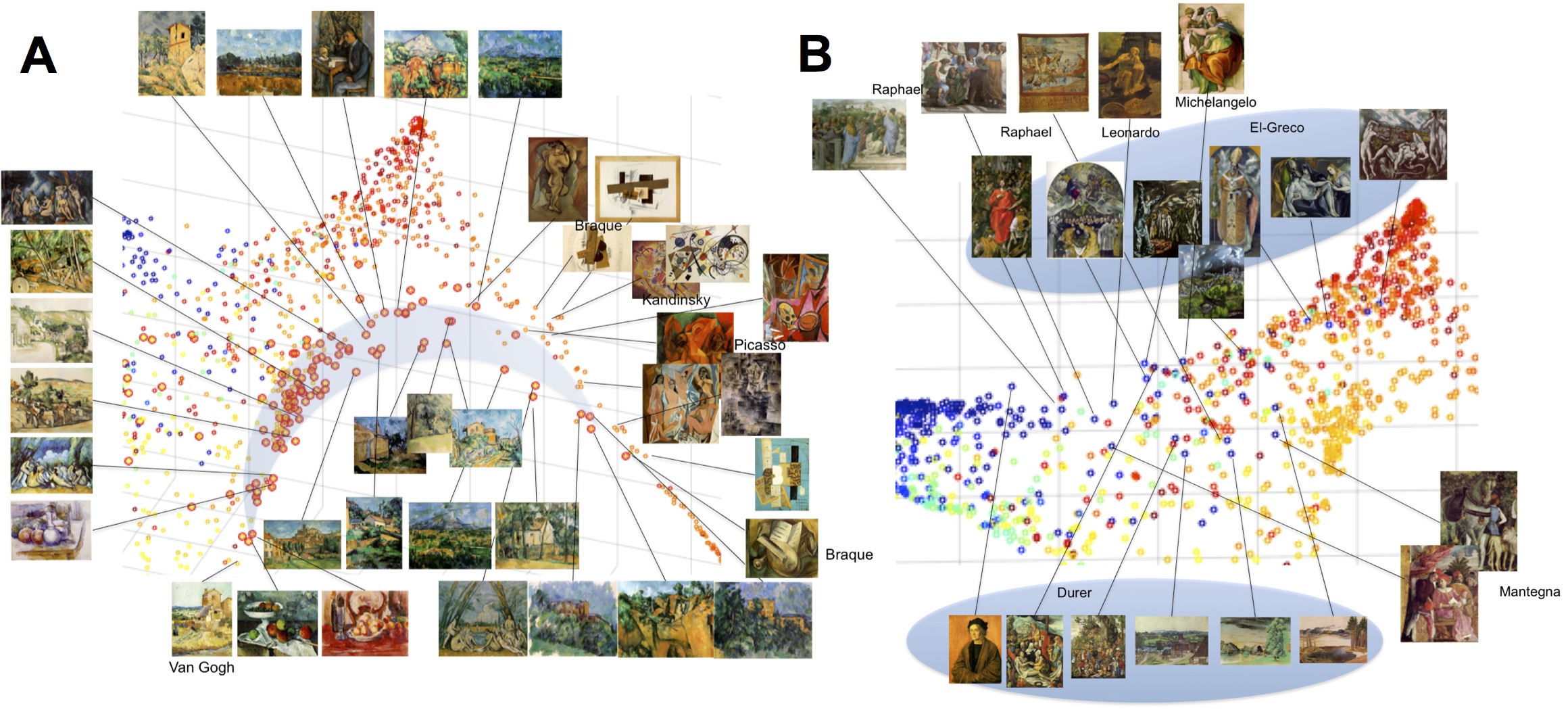}
\caption{Interesting connections. (A) C\'ezanne's bridge: we can see branching at Post-Impressionism where C\'ezanne's work clearly separates from the other Post-Impressionist and Expressionist works towards the top. This branch continues to evolve till it connects to early Cubist works by Picasso and Braque, as well as abstracts works by Kandinsky. All thumbnails without labels in this plot are by C\'ezanne. (B) The connection between Renaissance and modern styles is marked by the outliers in the temporal progress patterns by certain works by El-Greco, D\"urer, Raphael, Mantegna, and Michelangelo.}
\label{FBridges}
\end{figure*}

One interesting connection that is captured in this representation is the arrangement of C\'ezanne's work in a way that connects Impressionism to Cubism. We can clearly see in Figures~\ref{FLLE}-B and ~\ref{FBridges}-A that C\'ezanne's work acting as a bridge between Impressionism at one side and Cubism and Abstract at the other side. Art historians consider C\'ezanne to be a key figure in the style transition towards Cubism and the development of abstraction in the 20th century art. This bridge of C\'ezanne's painting in the learned representation is quite interesting because that is a quantifiable connection in the data, not just a metaphorical term. We can see branching at Post-Impressionism where C\'ezanne's work clearly separates from the other Post impressionist and expressionist works towards the top. This branch continues to evolve until it connects to early Cubist works by Picasso and Braque, as well as abstract works by Kandinsky.  

Figures ~\ref{FLLE}-B and ~\ref{FBridges}-B show another interesting connection between the Renaissance and modern art as captured by the learned representation. Despite the fact that the structure reflects smooth temporal progression, it is interesting to see outlier to this progression. In particular there are some High Renaissance, Northern Renaissance and Mannerist paintings that stick out of the Renaissance cluster to the left and connect to art from late 19th and early 20th centuries. This is because frequent similarity between art works across time results in pulling influential works of art out of order and placing them closer to the art they may have influenced. We can see in the figure that the works that stick out of the Renaissance cluster at the left and connect to modernity are mainly dominated by some paintings by El-Greco and some paintings by D\"urer. Among the paintings by El-Greco that significantly stick out are Laoco\"on, Saint Ildefonso, View of Toledo, and Piet\`a. We can also see works by Raphael, Mantegna, and Michelangelo in this group as well.  


We can accentuate this representation by the reducing the number of neighbors of each painting while constructing the manifold. This small neighborhood construction results in lining up distinctive paintings in thin structures to accentuate trends.  This results of this version shown in Figure~\ref{FLLE}-C\footnote{The accentuated representation is achieved by setting the neighborhood size to 25 to reduce the connectivity and accentuate the visualization.}. In this visualization, five distinct styles are clearly accentuated among all the twenty styles that the machine learned: Renaissance, Baroque, Impressionism, Cubism, and Abstract. We show the representative paintings at the tip of each of the five styles, which are in order Raphael's St. Catherine (1507), Rembrandt's self-portrait (1640), Monet's Water Lilies under clouds (1903), Braque's Violin and Candlestick (1910), and Malevich's Red Square (1915).  Both C\'ezanne's bridge and El-Greco/D\"urer's bridge are accentuated in this representation as well.

Interestingly, both C\'ezanne's connection and El-Greco/D\"urer connection appears consistently in the various representations learned by different networks, however manifested in different forms. Another interesting observation is that Cubism appears as very unique style that sticks out of as a singularity in most of the representations.


\section{Conclusions and Discussion}

In this paper we presented the result of a comprehensive study of training various deep convolutional neural networks for the purpose of style classification. We mainly emphasized analyzing the learned representation to get an understanding of how the machine achieves style classification. 

The different representations learned by the machine using different networks shared striking similarities in many aspects. It is clear that a small numbers of factors can explain most of the variance in the studied datasets in all the trained models (between 7 and 10 factors explains 95\% of the variance), which indicates that a small number of factors encapsulate the characterization of different styles in art history. 

The results also indicate that the learned representation shows a smooth temporal transition between styles, which the machine discovered without any notion of time given at training. Most interestingly, studying the modes of variations in the representation showed a radial temporal progress, with quantifiable correlation with time, starting at Renaissance, to Baroque, progressing all the way to Impressionism, Post impression, Cubism and closing the loop with 20th century styles such as Abstract and Pop Art coming back close to Renaissance. 
By studying the correlation between the modes of variations and W\"olfflin's suggested pairs, it was found that, consistently, all the learned models have the first mode of variation correlating the most with the concept of plane vs. recession while the second mode of variation correlates the most with the concept of linear vs. painterly. This correlation explains the radial temporal progress and the loop closure between Renaissance and 20th century styles since they share linearity and planarity in their form. 

We also studied the activation manifold of the different representation learned by the machine, which also reveal smooth temporal progress captured by the representation as well as correlation with W\"olfflin's concepts. Studying the activation manifolds with different neighborhood structure allowed us to discover different interesting connections in the history of art in a quantifiable way, such as the role of C\'ezanne's work as a bridge between a Impressionism and Cubism-Abstract art. Another interesting connection is the connection between Renaissance and modern styles such as Expressionism, Abstract-Expressionism, through the works of El-Greco, D\"urer, Raphael, and others.

Visualizing the different representations shows that certain artists were consistently picked by the machine as the distinctive representatives of the styles they belong to as they were the extreme points along the dimensions aligned with each style. For example, such distinctively representative artists are Van Eyck and D\"urer for Northern Renaissance, Raphael for Italian Renaissance, Rembrandt and Rubin for Baroque, Monet for Impressionism, C\'ezanne and Van Gogh for Post Impressionism, Rousseau for Naïve-Primitivism, Picasso and Braque for Cubism, and Malevich and Kandinsky for Abstract. While this is quite known for art historians, the machine discovered and highlighted these artists, amongst many others in each style, as the distinctive representatives of their styles without any prior knowledge and in a quantifiable way.

The networks are presented by raw colored images, and therefore, they have the ability to learn whatever features suitable to discriminate between styles, which might include compositional features, contrast between light and dark, color composition, color contrast, detailed brush strokes, subject matter related concepts. In particular, networks pre-trained on object categorization datasets might suggest potential bias towards choosing subject-matter-related features for classification. However, visualizing the learned representations reveals that the learned representations rule out subject matter as a basis for discrimination. This is clear from noticing the loop closure between Renaissance style, which is dominated with religious subject matter, and modern 20th century styles, such as Abstract, Pop art, and others. In contrast, this loop closure suggests that the basis of discrimination is related to concepts related to the form as suggested by W\"olfflin. 

The implication of the networks ability to recover a smooth temporal progression through the history of art, in absence of any temporal cues given at training, and in absence of any temporal constraints other than putting paintings of the same style closer to each other to achieve classification, suggests that visual similarity is the main factor that forces this smooth temporal representation to evolve, which in turns echoes the smooth transition of style in the history of art.

The results of this study highlight the potential role that data science and machine learning can play in the domain of art history by approaching art history as a predictive science to discover fundamental patterns and trends  not necessarily apparent to the individual human eye. The study also highlights the usefulness  of re-visiting the formal methods in art history pioneered by art historians such as W\"olfflin, in the age of data science using tools from computer vision and machine learning. Finally, the study offers insights into the characteristics and functions of style for art historians, confirming existing knowledge in an empirical way, and providing machine-produced patterns and connections for further exploration. 



\end{document}